\documentclass{article} 
\usepackage{iclr2016_workshop,times}

\usepackage{url}
\usepackage{dsfont}
\usepackage{algorithm}
\usepackage[noend]{algorithmic}
\usepackage{graphicx}
\usepackage{amssymb}
\usepackage{amsmath}

\renewcommand{\algorithmiccomment}[1]{\bgroup\hfill $\triangleright$ ~#1\egroup}

\usepackage{hyperref}
\title{CMA-ES for Hyperparameter Optimization of Deep Neural Networks}

\author{Ilya Loshchilov \& Frank Hutter \\
Univesity of Freiburg\\
Freiburg, Germany, \\
\texttt{\{ilya,fh\}@cs.uni-freiburg.de} 
}

\begin{document}

\maketitle

\begin{abstract} 
Hyperparameters of deep neural networks are often optimized by grid search, random search or Bayesian optimization. 
As an alternative, we propose to use the Covariance Matrix Adaptation Evolution Strategy (CMA-ES), which is known for its state-of-the-art performance in derivative-free optimization. CMA-ES has some useful invariance properties and is friendly to parallel evaluations of solutions. We provide a toy example comparing CMA-ES and state-of-the-art Bayesian optimization algorithms for tuning the hyperparameters of a convolutional neural network for the MNIST dataset on 30 GPUs in parallel. 
\end{abstract}


Hyperparameters of deep neural networks (DNNs) are often optimized by grid search, random search~\citep{bergstra-jmlr12a} or Bayesian optimization~\citep{snoek-nips12a,snoek2015scalable}. For the optimization of continuous hyperparameters, Bayesian optimization based on Gaussian processes~\citep{rasmussen-book06a} is known as the most effective method.
While for joint structure search and hyperparameter optimization, tree-based Bayesian optimization methods~\citep{hutter-lion11a,bergstra-nips11a} are known to perform better~\citep{bergstra-icml13a,eggensperger-bayesopt13,domhan2015speeding}, here we focus on continuous optimization. We note that integer parameters with rather wide ranges (e.g., number of filters) can, in practice, be considered to behave like continuous hyperparameters.

As the evaluation of a DNN hyperparameter setting requires fitting a model and evaluating its performance on validation data, this process can be very expensive, which often renders sequential hyperparameter optimization on a single computing unit infeasible. Unfortunately, Bayesian optimization is sequential by nature:
while a certain level of parallelization is easy to achieve by conditioning decisions on expectations over multiple hallucinated performance values for currently running hyperparameter evaluations~\citep{snoek-nips12a} or by evaluating the optima of multiple acquisition functions concurrently~\citep{hutter-lion12a,Chevalier2013,Desautels2014}, perfect parallelization appears {difficult to achieve} since the decisions in each step depend on all data points gathered so far.
Here, we study the use of a different type of derivative-free continuous optimization method where {parallelism is allowed by design}.

The Covariance Matrix Adaptation Evolution Strategy (CMA-ES \citep{hansen2001completely}) is a state-of-the-art optimizer for continuous black-box functions.
While Bayesian optimization methods often perform best for small function evaluation budgets (e.g., below 10 times the number of hyperparameters being optimized), CMA-ES tends to perform best for larger function evaluation budgets; for example, \citet{loshchilov2013bi} showed that CMA-ES performed best among more than 100 classic and modern optimizers on a wide range of blackbox functions. 
{CMA-ES has also been used for hyperparameter tuning before, e.g., for tuning its own Ranking SVM surrogate models \citep{loshchilov2012self} or for automatic speech recognition \citep{watanabe2014black}}.  

In a nutshell, CMA-ES is an iterative algorithm, that, in each of its iterations, samples $\lambda$ candidate solutions from a multivariate normal distribution, evaluates these solutions {(sequentially or in parallel)} and then adjusts the sampling distribution used for the next iteration to give higher probability to good samples. (Since space restrictions disallow a full description of CMA-ES, we refer to \citet{hansen2001completely} for details.)
Usual values for the so-called population size $\lambda$ are around 10 to 20; in the study we report here, we used a larger size $\lambda=30$ to take full benefit of 
30 GeForce GTX TITAN Black GPUs we had available. Larger values of $\lambda$ are also known to be helpful for noisy and multi-modal problems. 
Since all variables are scaled to be in [0,1],
we set the initial sampling distribution to $\mathcal{N}(0.5,0.2^2)$.
%
We didn't try to employ any noise reduction techniques \citep{hansen2009method} or surrogate models \citep{loshchilov2012self}. 

\begin{figure*}[t]
\begin{center}
\includegraphics[width=0.45\textwidth]{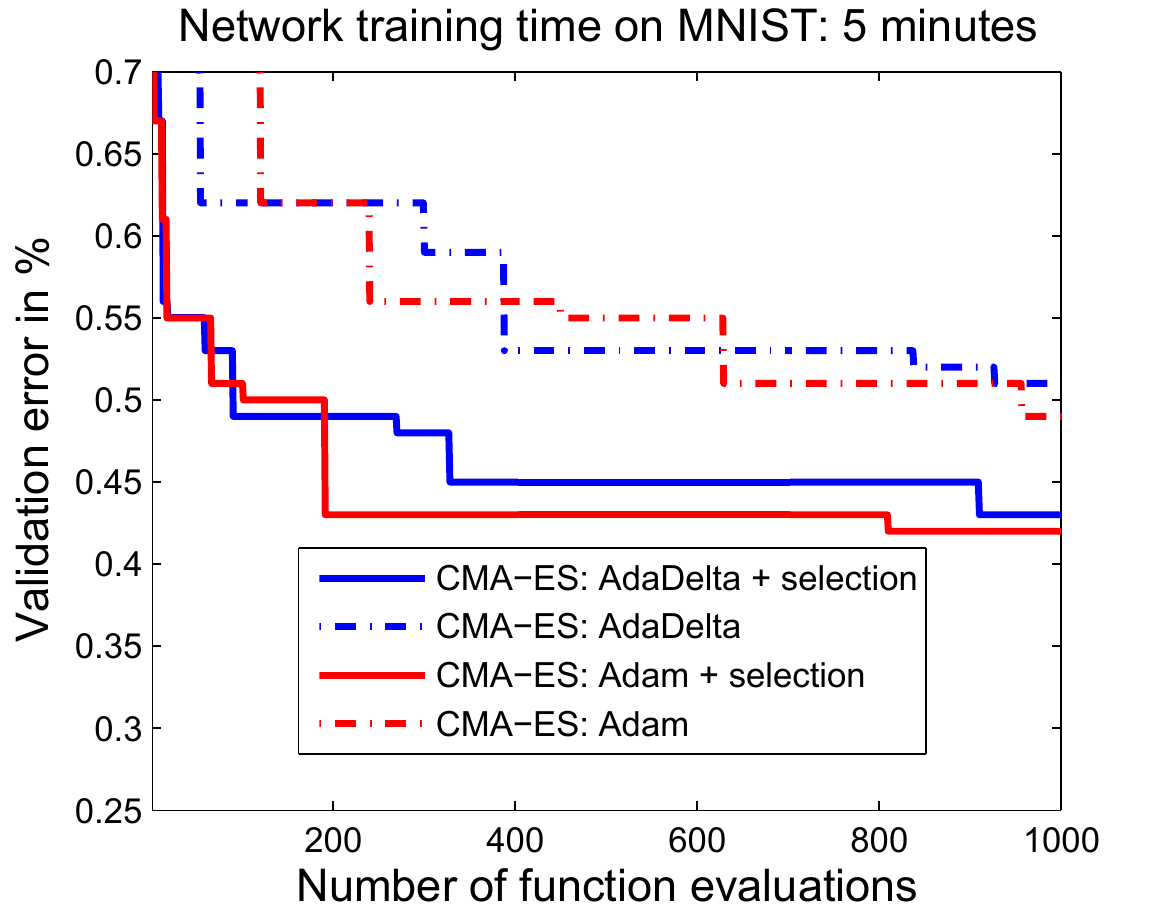}
\includegraphics[width=0.45\textwidth]{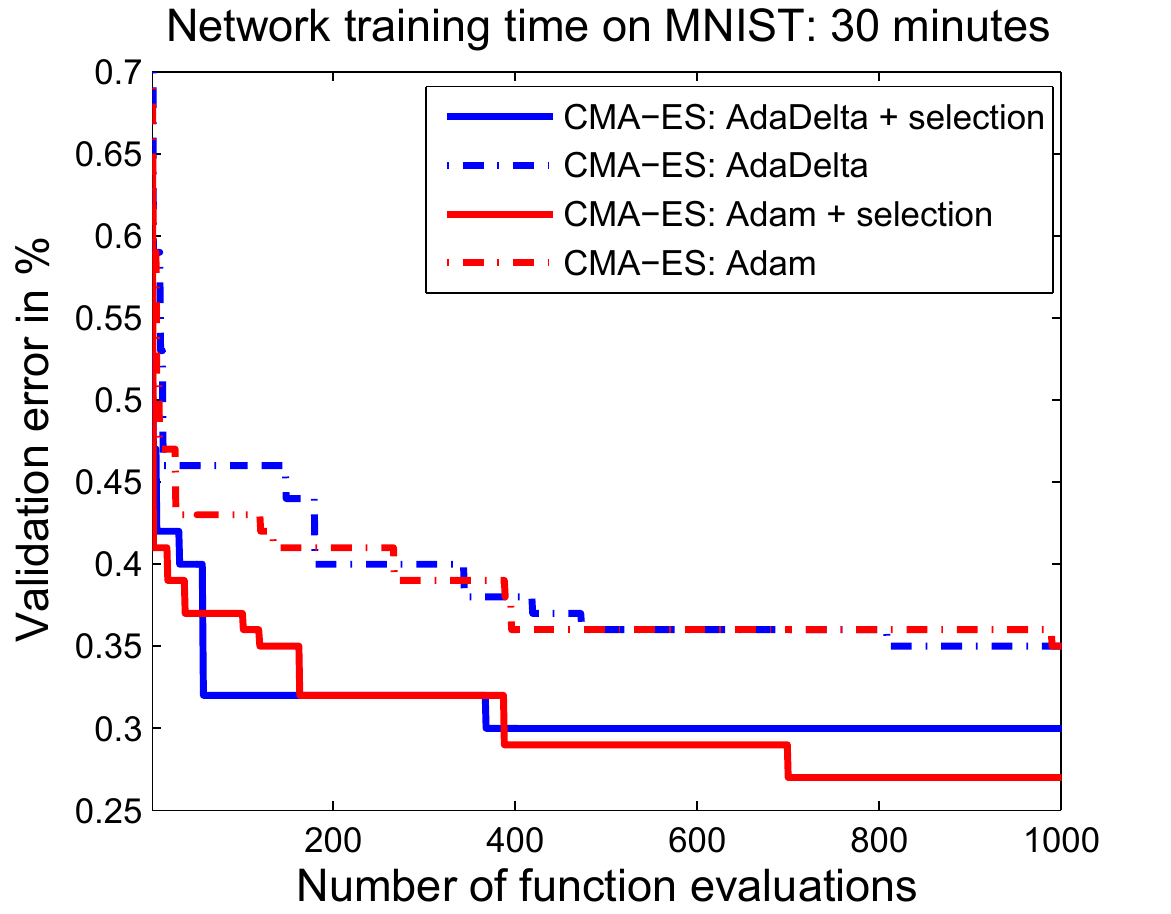}\\
\end{center}
\caption{
Best validation errors CMA-ES found for AdaDelta and Adam with and without batch selection 
when hyperparameters are optimized by CMA-ES with training time budgets of  
5 minutes (left) and 30 minutes (right).
}
\label{FigureMnist_CMA}
\vspace*{-0.4cm}
\end{figure*}

In the study we report here, we used AdaDelta \citep{zeiler2012adadelta} and Adam \citep{kingma2014adam} to train DNNs on the MNIST dataset (50k original training and 10k original validation examples). The 19 hyperparameters describing the network structure and the learning algorithms are given in Table~\ref{table:table}; the code is also available at \url{https://sites.google.com/site/cmaesfordnn/} (anonymous for the reviewers). 
We considered both the default (shuffling) and online loss-based batch selection of training examples~\citep{loshchilov2015online}. 
The objective function is the smallest validation error found in all epochs when the training time (including the time spent on model building) is limited. 
Figure \ref{FigureMnist_CMA} shows the results of running CMA-ES on 30 GPUs on eight different hyperparameter optimization problems: all combinations of using (1)~AdaDelta \citep{zeiler2012adadelta} or Adam \citep{kingma2014adam}; (2)~standard shuffling batch selection or batch selection based on the latest known loss \citep{loshchilov2015online}; and (3)~allowing 5 minutes or 30 minutes of network training time. We note that in all cases CMA-ES steadily improved the best validation error over time and in the best case yielded validation errors below 0.3\% in a network trained for only 30 minutes (and 0.42\% for a network trained for only 5 minutes).
We also note that batch selection based on the latest known loss performed better than shuffling batch selection and that the results of AdaDelta and Adam were almost indistinguishable. Therefore, the rest of the paper discusses only the case of Adam with batch selection based on the latest known loss.

We compared the performance of CMA-ES against various state-of-the-art Bayesian optimization methods.  
The main baseline is GP-based Bayesian optimization, as implemented by the widely known Spearmint system~\citep{snoek-nips12a} (available at \url{https://github.com/HIPS/Spearmint}).
In particular, we compared to Spearmint with two different acquisition functions:
(i)~Expected Improvement (EI), as described by \cite{snoek2012practical} and implemented in the main branch of Spearmint; and 
(ii)~Predictive Entropy Search (PES), as described by \cite{hernandez2014predictive} and implemented in a sub-branch of Spearmint (available at \url{https://
github.com/HIPS/Spearmint/tree/PESC}). 
Experiments by \cite{hernandez2014predictive} demonstrated that PES is superior to EI; our own (unpublished) preliminary experiments on the black-box benchmarks used for the evaluation of CMA-ES by \cite{loshchilov2013bi} also confirmed this. Both EI and PES 
have an option to notify the method about whether the problem at hand is noisy or noiseless. To avoid a poor choice on our side, we ran both algorithms in both regimes. Similarly to CMA-ES, in the parallel setting we set the maximum number of concurrent jobs in Spearmint to 30. 
{We also benchmarked the tree-based Bayesian optimization algorithms TPE~\citep{bergstra-nips11a} and SMAC \citep{hutter-lion11a} (with 30 parallel workers each in the parallel setting). TPE accepts prior distributions for each parameter, and we used the same priors $\mathcal{N}(0.5,0.2^2)$ as for CMA-ES.}

\begin{figure*}[t]
\begin{center}
\includegraphics[width=0.495\textwidth]{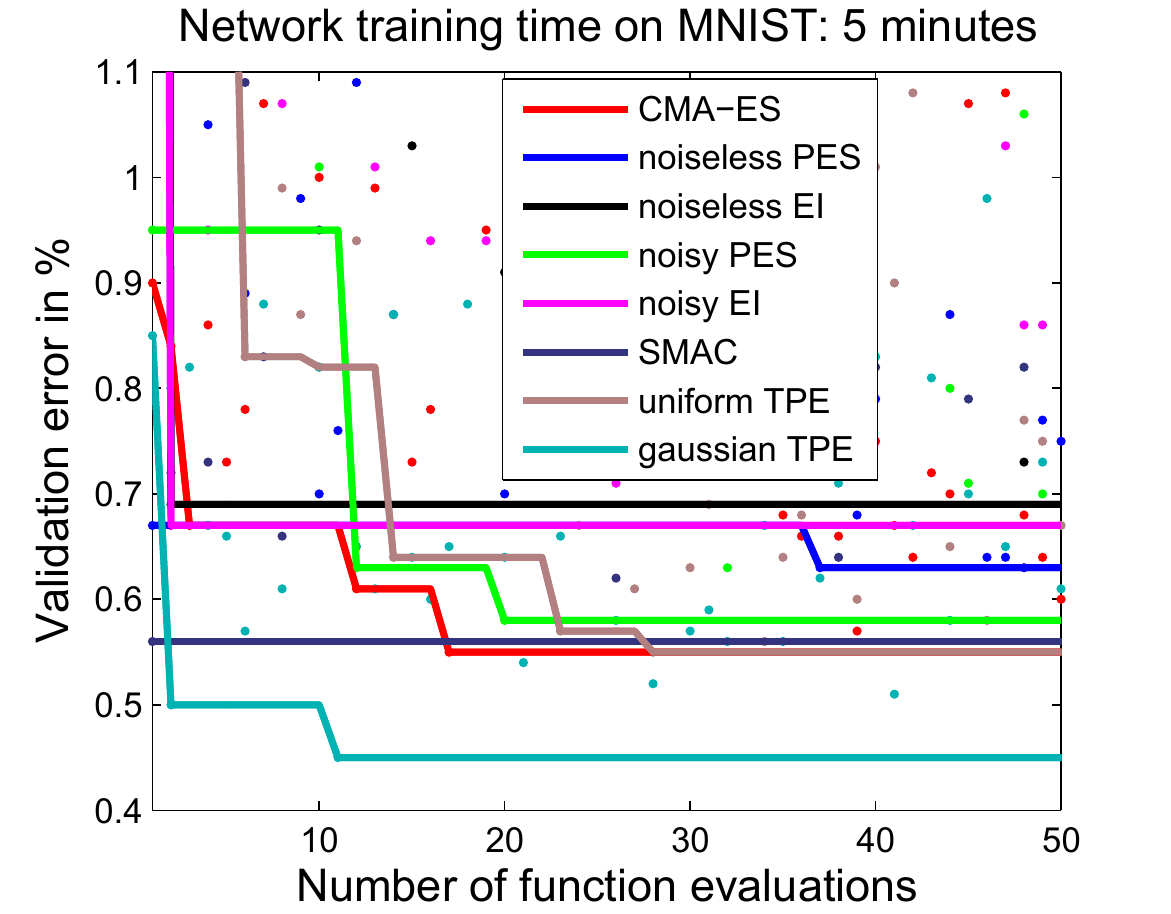}
\includegraphics[width=0.495\textwidth]{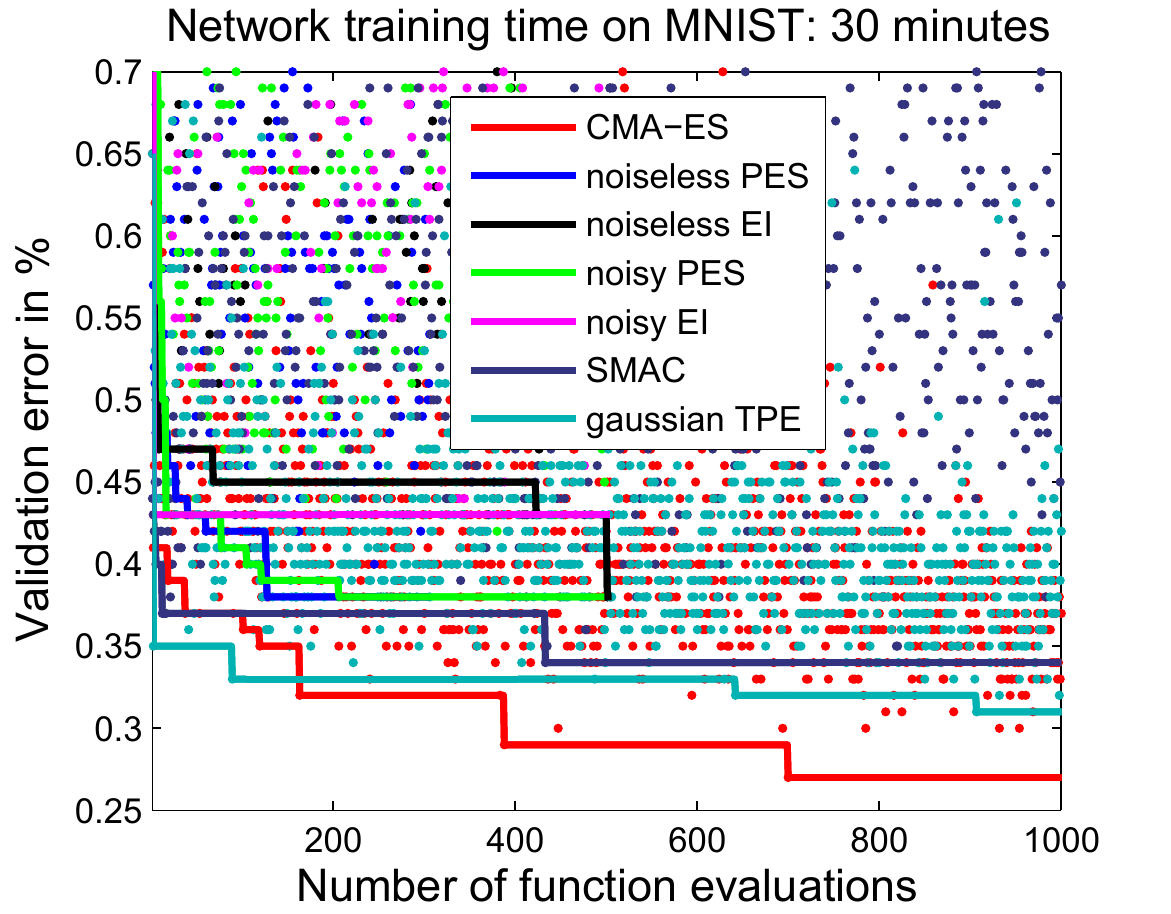}
\end{center}
\caption{
Comparison of optimizers for Adam with batch selection when solutions are evaluated sequentially for 5 minutes each (left), and in parallel for 30 minutes each (right).
Note that the red dots for CMA-ES were plotted first and are in the background of the figure (see also Figure \ref{FigureMNISThist} in the supplementary material for an alternative representation of the results).
}
\label{FigureMnist_all}
\end{figure*}

Figure \ref{FigureMnist_all} compares the results of CMA-ES vs.{} Bayesian optimization with EI\&PES, {SMAC and TPE}, both in the sequential and in the parallel setting. In this figure, to illustrate the actual function evaluations, each evaluation within the range of the y-axis is depicted by a dot. Figure \ref{FigureMnist_all}~(left) shows the results of all tested algorithms when solutions are evaluated sequentially with a relatively small network training time of 5 minutes each. 
Note that we use CMA-ES with $\lambda=30$ and thus the first 30 solutions are sampled from the prior isotropic (not yet adapted) Gaussian with a mean of 0.5 and standard deviation of 0.2. Apparently, the results of this sampling are as good as the ones produced by EI\&PES. 
This might be because of a bias towards the middle of the range, 
or because EI\&PES do not work well on this noisy high-dimensional problem, or because of both. Quite in line with the conclusion of \cite{bergstra-jmlr12a}, it seems that the presence of noise and rather wide search ranges of hyperparameters make sequential optimization \textit{with small budgets} rather inefficient {(except for TPE)}, i.e., as efficient as random sampling. 
{SMAC started from solutions in the middle of the search space and thus performed better than the Spearmint versions, but it did not improve further over the course of the search. TPE with Gaussian priors showed the best performance.}

Figure \ref{FigureMnist_all}~(right) shows the results of all tested algorithms when solutions are evaluated \textit{in parallel} on 30 GPUs.
Each DNN now trained for 30 minutes, meaning that, for each optimizer, running this experiment sequentially would take 30\,000 minutes (or close to 21 days) on one GPU; in parallel on 30 GPUs, it only required 17 hours. Compared to the sequential 5-minute setting, the greater budget of the parallel setting allowed CMA-ES to improve results such that most of its latest solutions had validation error below 0.4\%. 
The internal cost of CMA-ES was virtually zero, but it was a substantial factor for EI\&PES due to the cubic complexity of standard GP-based Bayesian optimization: after having evaluated 100 configurations, it took roughly 30 minutes to generate 30 new configurations to evaluate, and as a consequence 
500 evaluations by EI\&PES took more wall-clock time than 1000 evaluations by CMA-ES. 
This problem could be addressed by 
using approximate GPs~\cite{rasmussen-book06a} 
or another efficient multi-core implementation of Bayesian Optimization, such as the one by \cite{snoek2015scalable}. However, the Spearmint variants also performed poorly compared to the other methods in terms of validation error achieved. One reason might be that this benchmark was too noisy and high-dimensional for it. TPE with Gaussian priors showed good performance, which was dominated only by CMA-ES after about 200 function evaluations.

Importantly, the best solutions found by TPE with Gaussian priors and CMA-ES often coincided and typically do not lie in the middle of the search range (see, e.g., $x_3, x_6, x_9, x_{12}, x_{13}, x_{18}$ in Figure \ref{xxmnist} of the supplementary material).

In conclusion, we propose to consider CMA-ES as one alternative in the mix of methods for hyperparameter optimization of DNNs. It is powerful, computationally cheap and natively supports parallel evaluations. Our preliminary results suggest that CMA-ES can be competitive especially in the regime of parallel evaluations. However, we still need to carry out a much broader and more detailed comparison, involving more test problems and various modifications of the algorithms considered here, such as the addition of constraints~\citep{Gelbart2014,hernandez2014predictive}.

\small
\bibliography{iclr2016_conference}
\bibliographystyle{iclr2016_conference}

\cleardoublepage

\section{Supplementary Material}

\begin{figure*}[ht]
\begin{center}
\includegraphics[width=0.99\textwidth]{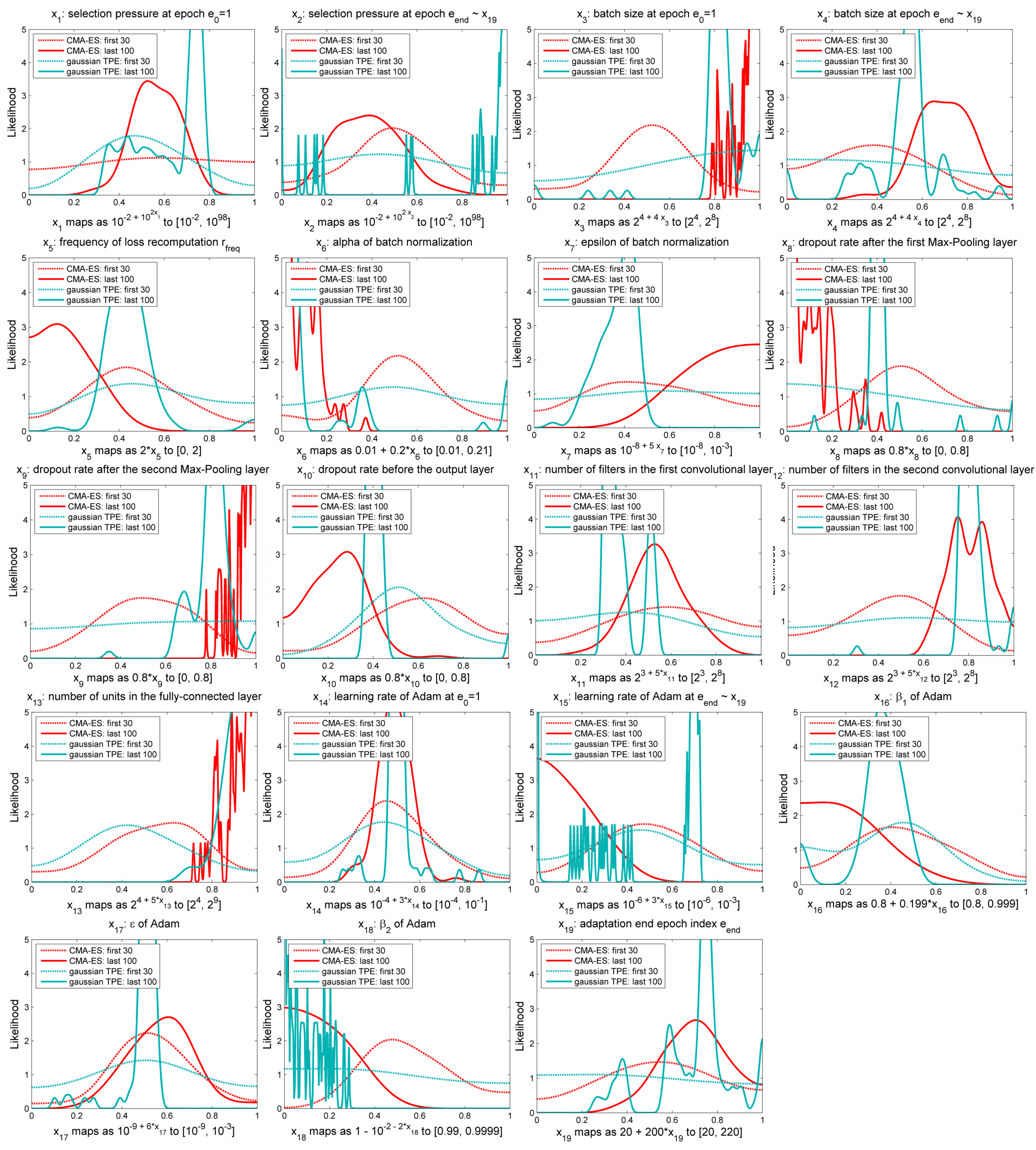}
\end{center}
\caption{
{Likelihoods of hyperparameter values to appear in the first 30 evaluations (dotted lines) and last 100 evaluations (bold lines) out of 1000 for CMA-ES and TPE with Gaussian priors during hyperparameter optimization on the MNIST dataset. We used kernel density estimation  via diffusion by \cite{botev2010kernel} with 256 mesh points.}    
}
\label{xxmnist}
\end{figure*}

\begin{figure*}[t]
\begin{center}
\includegraphics[width=0.995\textwidth]{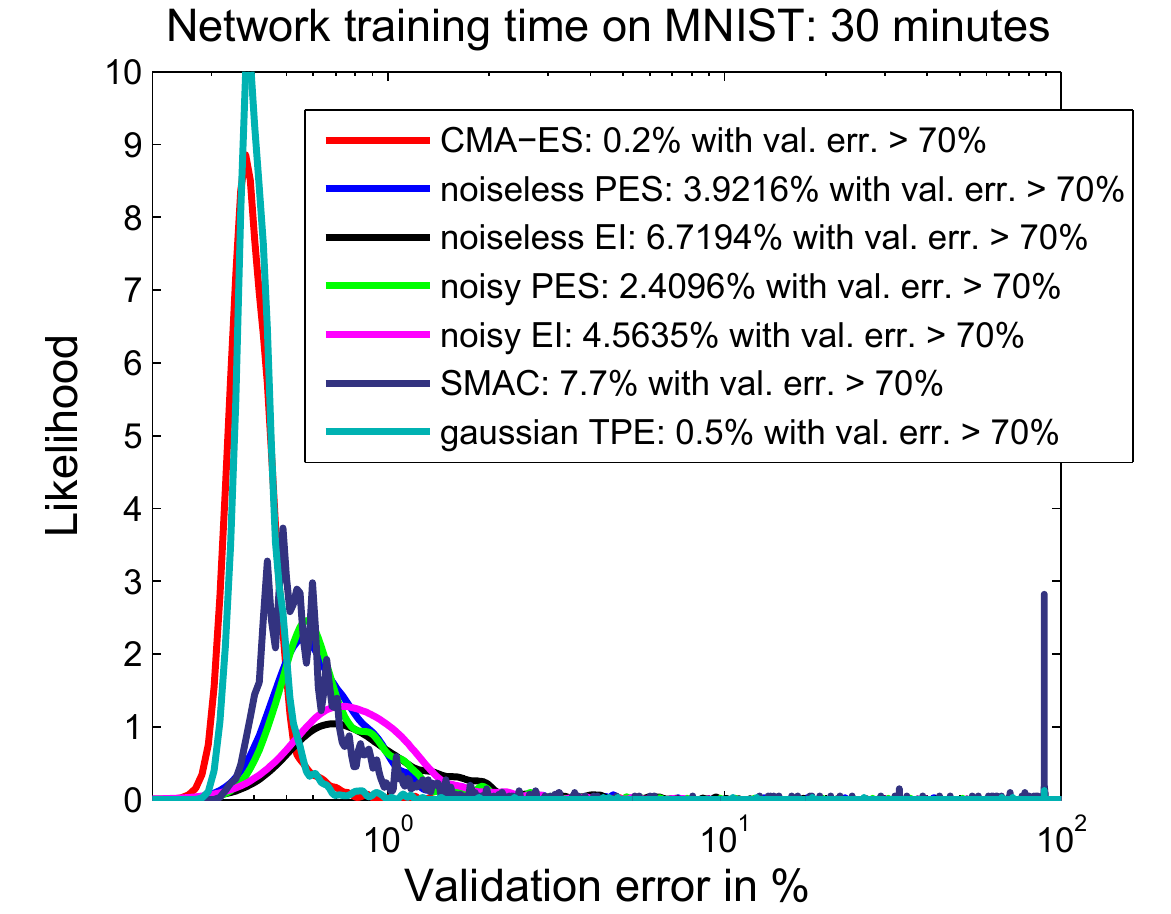}
\end{center}
\caption{
{Likelihoods of validation errors on MNIST found by different algorithms as estimated from all evaluated solutions with the kernel density estimator by \cite{botev2010kernel} with 5000 mesh points. Since the estimator does not fit well the outliers in the region of about 90\% error, we additionally supply the information about the percentage of the cases when the validation error was greater than 70\% (i.e., divergence or close to divergence results), see the legend. }
}
\label{FigureMNISThist}
\end{figure*}

{
 \begin{table}[ht]
\small
\caption{Hyperparameters descriptions, pseudocode transformations and ranges {for MNIST experiments}}. 
\centering
\begin{tabular}{l*{6}{l}r}
name   	& description & transformation & range  \\
\hline
$x_1$ 	& selection pressure at ${e_0}$ & $10^{-2 + 10^{2 x_1}}$   & $[10^{-2}, 10^{98}]$  \\ 
$x_2$ 	& selection pressure at ${e_{end}}$ & $10^{-2 + 10^{2 x_2}}$   & $[10^{-2}, 10^{98}]$  \\
$x_3$ 	& batch size at ${e_{0}}$ & $2^{4 + 4 x_3}$   & $[2^4, 2^8]$  \\
$x_4$ 	& batch size at ${e_{end}}$ & $2^{4 + 4 x_4}$   & $[2^4, 2^8]$  \\
$x_5$ 	& frequency of loss recomputation $r_{freq}$ & $2 x_5$   & $[0, 2]$  \\
$x_6$ 	& alpha for batch normalization & $0.01 + 0.2 x_6$   & $[0.01, 0.21]$  \\
$x_7$ 	& epsilon for batch normalization & $10^{-8 + 5 x_7}$  & $[10^{-8}, 10^{-3}]$  \\
$x_8$ 	& dropout rate after the first Max-Pooling layer & $0.8 x_8$  & $[0, 0.8]$  \\
$x_9$ 	& dropout rate after the second Max-Pooling layer & $0.8 x_9$  & $[0, 0.8]$  \\
$x_{10}$ 	& dropout rate before the output layer & $0.8 x_{10}$  & $[0, 0.8]$  \\
$x_{11}$ 	& number of filters in the first convolution layer & $2^{3 + 5 x_{11}}$  & $[2^3, 2^8]$  \\
$x_{12}$ 	& number of filters in the second convolution layer & $2^{3 + 5 x_{12}}$  & $[2^3, 2^8]$  \\
$x_{13}$ 	& number of units in the fully-connected layer & $2^{4 + 5 x_{13}}$  & $[2^4, 2^9]$  \\
$x_{14}$ 	& Adadelta: learning rate at ${e_{0}}$ & $10^{0.5 - 2 x_{14}}$  & $[10^{-1.5}, 10^{0.5}]$  \\
$x_{15}$ 	& Adadelta: learning rate at ${e_{end}}$ & $10^{0.5 - 2 x_{15}}$  & $[10^{-1.5}, 10^{0.5}]$  \\
$x_{16}$ 	& Adadelta: $\rho$ & $0.8 + 0.199 x_{16}$  & $[0.8, 0.999]$  \\
$x_{17}$ 	& Adadelta: $\epsilon$ & $10^{-3 - 6 x_{17}}$  & $[10^{-9}, 10^{-3}]$  \\
$x_{14}$ 	& Adam: learning rate at ${e_{0}}$ & $10^{-1 - 3 x_{14}}$  & $[10^{-4}, 10^{-1}]$  \\
$x_{15}$ 	& Adam: learning rate at ${e_{end}}$ & $10^{-3 - 3 x_{15}}$  & $[10^{-6}, 10^{-3}]$  \\
$x_{16}$ 	& Adam: $\beta_1$ & $0.8 + 0.199 x_{16}$  & $[0.8, 0.999]$  \\
$x_{17}$ 	& Adam: $\epsilon$ & $10^{-3 - 6 x_{17}}$  & $[10^{-9}, 10^{-3}]$  \\
$x_{18}$ 	& Adam: $\beta_2$ & $1 - 10^{-2 - 2 x_{18}}$  & $[0.99, 0.9999]$  \\
$x_{19}$ 	& adaptation end epoch index ${e_{end}}$ & $20 + 200 x_{19}$  & $[20, 220]$  \\
\end{tabular}
\label{table:table} 
\end{table}
}

\begin{figure*}[t]
\begin{center}
\includegraphics[width=0.495\textwidth]{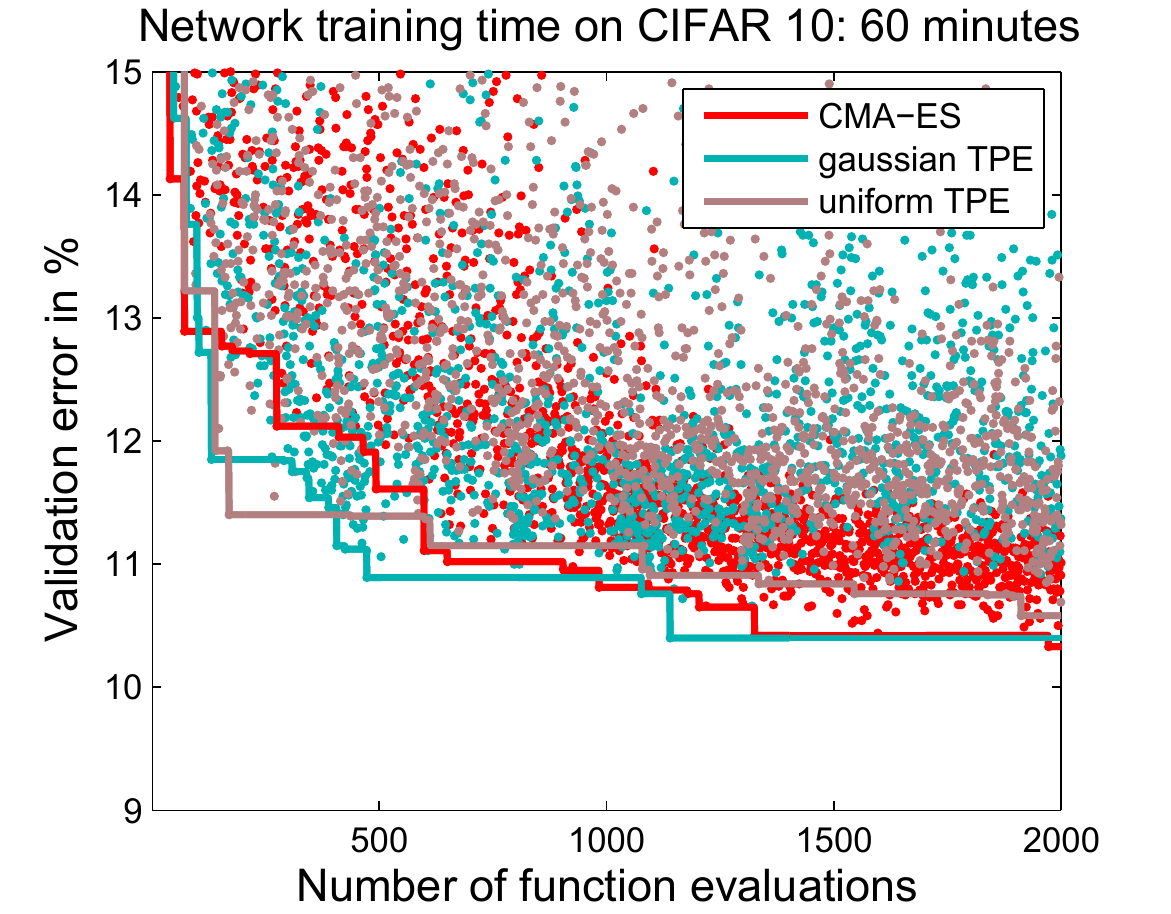}
\includegraphics[width=0.495\textwidth]{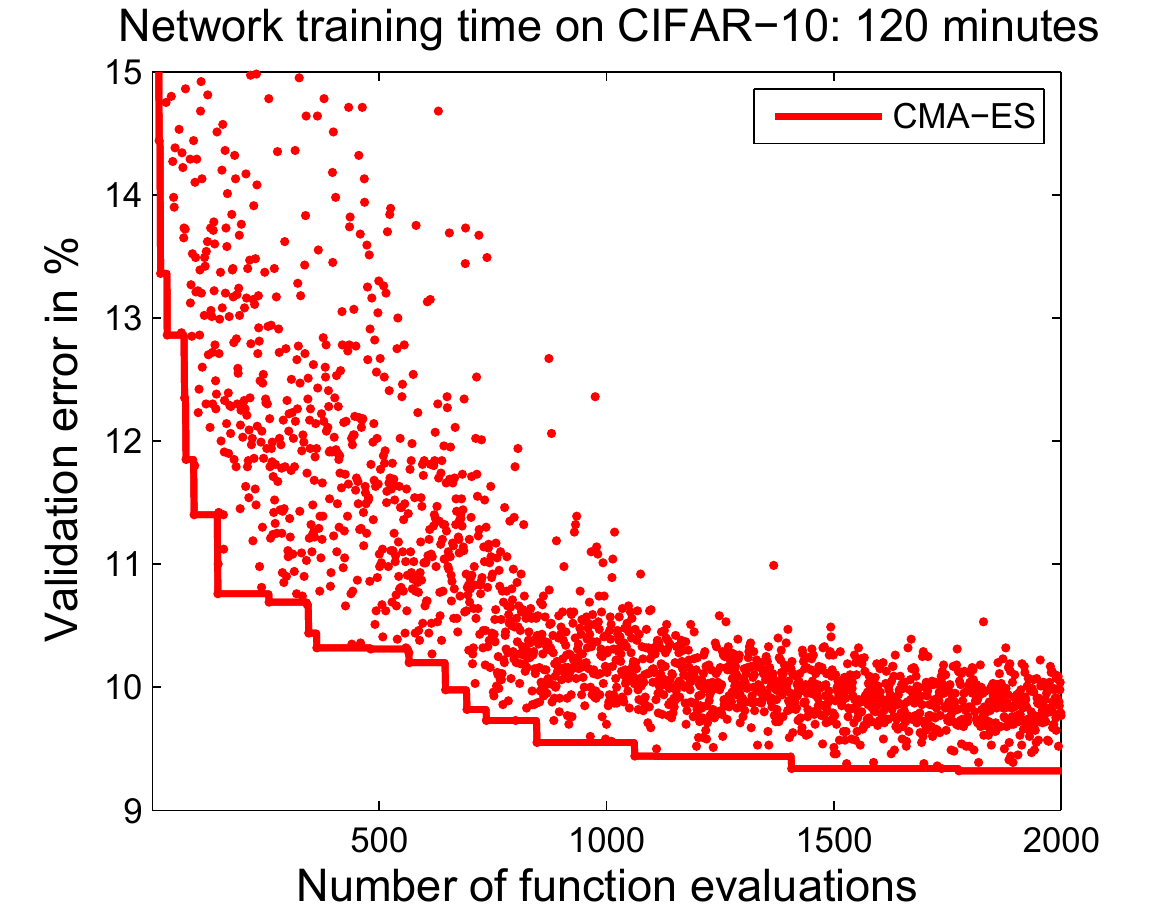}
\end{center}
\caption{
{Preliminary results not discussed in the main paper. Validation errors on CIFAR-10 found by Adam when hyperparameters are optimized by CMA-ES and TPE with Gaussian priors with training time budgets of 60 and 120 minutes. No data augmentation is used, only ZCA whitening is applied. Hyperparameter ranges are different from the ones given in Table 1 as the structure of the network is different, it is deeper.} 
}
\label{FigureCifar}
\end{figure*}

\end{document}